\documentclass[1p,sort&compress,number,12pt]{elsarticle}

\usepackage[T1]{fontenc}
\usepackage[utf8]{inputenc}
\usepackage{lmodern}
\usepackage[english]{babel}
\usepackage{enumitem}
\usepackage{amsmath}
\usepackage{amssymb}
\usepackage{graphicx}
\usepackage{wrapfig}
\usepackage{pdflscape}
\usepackage{float}
\usepackage{xcolor}
\usepackage{algorithm}
\usepackage{algorithmic}
\usepackage{multirow}
\usepackage{bbm}

\usepackage{fancyhdr}
\begin{document}

\begin{frontmatter}
\title{Unsupervised Feature Selection Based on the Morisita Estimator of Intrinsic Dimension}
\author{Jean GOLAY and Mikhail KANEVSKI}
\address{Institute of Earth Surface Dynamics, Faculty of Geosciences and Environment, University of Lausanne, 1015 Lausanne, Switzerland. Email: jean.golay@unil.ch.}

\begin{abstract}
This paper deals with a new filter algorithm for selecting the smallest subset of features carrying all the information content of a data set (i.e. for removing redundant features). It is an advanced version of the fractal dimension reduction technique, and it relies on the recently introduced Morisita estimator of Intrinsic Dimension (ID). Here, the ID is used to quantify dependencies between subsets of features, which allows the effective processing of highly non-linear data. The proposed algorithm is successfully tested on simulated and real world case studies. Different levels of sample size and noise are examined along with the variability of the results. In addition, a comprehensive procedure based on random forests shows that the data dimensionality is significantly reduced by the algorithm without loss of relevant information. And finally, comparisons with benchmark feature selection techniques demonstrate the promising performance of this new filter.
\end{abstract}

\end{frontmatter}

%main text
\section{Introduction}
Recent breakthroughs in technology have radically improved our ability to collect and store data. Consequently, more and more variables (or features \footnote{In this paper, the term ``feature'' is used as a synonym for ``variable''.}) are available to perform data mining tasks, but in general, a lot of them are redundant (i.e. they do not carry additional information beyond that subsumed by other features), or partially redundant, and contribute to the emergence of four major issues: (1) the reduction in the accuracy of learning algorithms because of the curse of dimensionality \cite{Bell61}, (2) the computer performance limitations related to memory and processing speed, (3) the difficulty in visualizing large amounts of complex and high-dimensional data and (4) the interpretability of the results which becomes less tractable making it difficult to gain an insight into the mechanisms that generated the data. 

Due mainly to these redundant and partially redundant features, data points do not occupy the full $E$-dimensional space $\mathbb{R}^E$ ($E$ is the number of features in a data set) in which they are embedded. Instead, they are often regarded as residing on a lower $M$-dimensional manifold where $M (\leqslant E)$ is the \textit{Intrinsic Dimension} (ID) of data \cite{Cama16}. Dimensionality Reduction (DR) methods \cite{LeeVer07,Liu15} can help remove redundant information by trying to map the original data space coordinates to an intrinsic coordinate system of dimensionality $M$. Depending on the assumptions made about the shape of the manifold, the mapping can be either linear (e.g. PCA \cite{Pear01}) or non-linear (e.g. kernel-PCA \cite{Schol99}), and a great advantage of the DR approach is its potential to capture complex dependencies. On the other hand, DR often leads to a deterioration in the physical interpretability of the data and to difficulties in the understanding of subsequent results. A possible solution to these drawbacks is the implementation of feature selection methods.

The goal of \textit{feature selection} \cite{Guy06,Liu07,Koh97,Fen14,Pen05} is to select the smallest subset of original features which maintains some meaningful characteristics with respect to a chosen criterion. According to the possible use of output information (e.g. class labels), feature selection methods can be broadly classified as either supervised or unsupervised. Advanced supervised methods aim to select features which are both relevant to the prediction (i.e. classification or regression) of some output information and related as little as possible to one another (i.e. select relevant and non-redundant features). In contrast, unsupervised methods do not make use of any a priori knowledge regarding an output, and they can be further divided into two categories: Cluster Recognition (CR) and Redundancy Minimization (RM).

The CR methods aim to find the smallest subset of features that uncovers the most ``interesting'' and ``natural'' groupings (i.e. clusters) of data points \cite{Dy04,Dy07,Ale13,Chen15}. They rely on criteria of relevance that do not involve any output information, and they can be categorized into filters and wrappers \cite{Joh94}. The former (e.g. the Laplacian score method \cite{He06}, SPEC \cite{Zha07,Zha11} and MCFS \cite{Cai10}) do not incorporate the clustering algorithm that will ultimately be applied, while the latter do (e.g. methods introduced in \cite{Dy04,Dy03} or reviewed in \cite{Dy07}). In contrast, the RM methods are often not restricted to clustering problems, and they can be used as preprocessing tools in a wide variety of data mining approaches. Their goal is to select the smallest subset of features in such a way that all the information content of a data set is preserved as much as possible. In other words, they aim to eliminate all the redundant information by selecting the most informative features (i.e. the non-redundant features). To achieve this goal, the RM methods often use criteria based on PCA loading values \cite{Kim11} or on measures of feature dependency, such as the maximal information compression index \cite{Mit02}, mutual information \cite{Fen16,Mart07} and fractal-based measures of ID \cite{Trai00,Sousa07,Mo12}. More recently, Wang et al. \cite{Wan15} proposed a criterion that minimizes the reconstruction error of a linear projection of the original features, while ensuring low redundancy. Further, the RM methods can be thought of as filters, and like many other methods of feature selection, they can rely on greedy (e.g. Sequential Forward Selection (SFS) \cite{Whit71} and Sequential Backward Elimination (SBE) \cite{Ree99,Har98}) or randomized (e.g. simulated annealing \cite{Kirk83}) search strategies if they consider multivariate interactions and aim to find the best subset among the $2^E-1$ combinations of features. Lastly, methods combining the CR and the RM approaches have also been developed. Many of them use a graph Laplacian matrix to preserve the data structure and involve a low redundancy constraint or a more advanced regularization term \cite{Han15}.

More specifically, the use of ID for unsupervised feature selection was introduced by Traina et al. \cite{Trai00,Trai10}. They extended the concept of ID to fractal dimensions and proposed the Fractal Dimension Reduction (FDR) algorithm. FDR is a filter algorithm for non-linear RM that follows a SBE search strategy. It aims to eliminate the features which do not contribute to increasing the value of the data ID (i.e. the ID of the studied data set), and it relies on R\'{e}nyi's dimension of order $2$ \cite{Hent83}, $D_2$, for the ID estimation. An extension to FDR was proposed by De Sousa et al. \cite{Sousa07} to identify subsets of correlated attributes in databases according to user-defined levels of correlation. Finally, Mo and Huang \cite{Mo12} modified FDR by replacing $D_2$ with the correlation dimension $df_{cor}$ \cite{Grass833}.

The present paper deals with a novel ID-based filter algorithm for RM. It relies on the recently introduced Morisita estimator of ID, $M_m$, which was shown to be more effective than $D_2$ and $df_{cor}$ in situations where the data points were sparsely distributed \cite{Go15}. Besides, the proposed algorithm follows a SFS search strategy; it can process large and highly non-linear data, and its implementation is straightforward in \textsf{R} and \textsc{Matlab}. Another advantage is that the number of features to be selected can be determined directly from the results. And it is also worth mentioning that $M_m$ was already used successfully to perform supervised feature selection in regression problems \cite{Go15Esann}. 

The remainder of this paper is organized as follows. Section \ref{Mindex} presents the Morisita estimator of ID, and Section \ref{ID_cor} explains the relationship between ID and data redundancy. In Section \ref{ID_RM}, the proposed algorithm for RM is introduced, and Section \ref{exp_res} is devoted to numerical experiments conducted on simulated data and on real world case studies from the UCI machine learning repository. The quality of the results is assessed using a comprehensive methodology based on random forests \cite{Bre01}, and comparisons with benchmark feature selection techniques (including FDR) are also discussed. Finally, conclusions are drawn in the last section with a special emphasis on potentialities and future challenges.

\begin{figure}
\centering
\includegraphics[width=\linewidth]{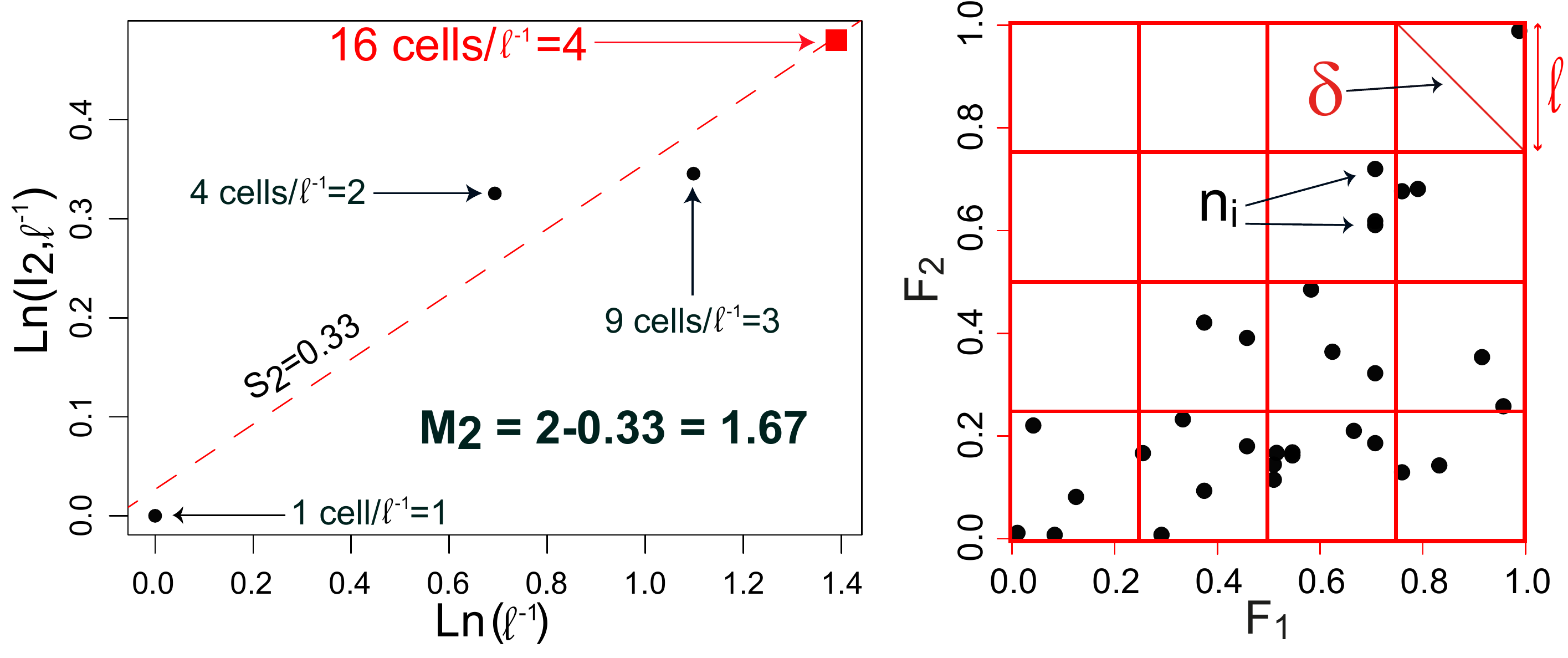}
\caption{The Morisita estimator of ID, $M_2$, applied to two features of the \textsf{R} data set ``Trees'' \cite{Rcran}. The red square in the left-hand panel indicates the ln value of the multipoint Morisita index computed with the grid displayed on the right.}\label{Fig_Mindid}
\end{figure}

\section{The Morisita Estimator of Intrinsic Dimension}\label{Mindex}

\subsection{Overview}
The Morisita estimator of ID \cite{Go15}, $M_m$, is derived from the multipoint Morisita index $I_{m,\delta}$ \cite{Go14,Hul90,Mori59}. $I_{m,\delta}$ is computed by means of an $E$-dimensional grid of $Q$ cells (or quadrats) of diagonal size $\delta$ superimposed over the data points (see Figure \ref{Fig_Mindid}). It measures how many times more likely it is that $m$ ($m\geq 2$) points selected at random will be from the same cell than it would be if the $N$ points of the studied data set were distributed according to a random distribution generated from a Poisson process (i.e. complete spatial randomness). $I_{m,\delta}$ is given by the following formula: \begin{equation}\label{Eq_mindex}
I_{m,\delta}=Q^{m-1}\frac{\sum_{i=1}^Q n_i(n_i-1)(n_i-2) \dotsm (n_i-m+1)}{N(N-1)(N-2) \dotsm (N-m+1)}
\end{equation}where $n_i$ is the number of data points in the $i^{th}$ cell. In general, $m$ is set to $2$, and the computation of the index is iterated for $R$ different values of $\delta$. These values must be chosen by the user and determine the scales at which the phenomenon will be characterized. Within the range of these scale values, if the data set follows a fractal behaviour (i.e. is self-similar), the functional relationship between $\log{(I_{m,\delta})}$ and $\log{(1/\delta)}$ is linear, its slope, $S_m$, is the Morisita slope, and $M_m$ can be written as:\begin{equation}\label{Eq_mindid}M_m = E - \left( \frac{S_{m}}{m-1}\right).\end{equation}In practice, each feature is rescaled to the $[0,1]$ interval (so is the grid), and $\delta$ is replaced with the edge length, $\ell$, of the cells. In this context, $\ell^{-1}$ is simply the number of cells along each axis of the $E$-dimensional space where the data points are embedded.

\subsection{Detailed Procedure}
In the remainder of this paper, the Morisita estimator of ID will be used only with $m=2$ as advocated in \cite{Go15}. The following steps summarize how to compute the ID of a data set using $M_{m=2}$:
\begin{enumerate}[noitemsep]
\item Rescale each of the $E$ features to the $[0,1]$ interval.
\item \label{item_par} Choose the values of the parameter $\ell^{-1}$ so that the functional relationship of Step \ref{item_lm} can be well approximated by a linear regression model (see Subsection \ref{sub_par}).
\item Superimpose an $E$-dimensional grid over the data points. The size of the grid cells is controlled by the user through the parameter $\ell^{-1}$ which is simply the number of cells along each axis of the grid. 
\item Count the number of data points falling into the cells of the grid. This step must be repeated for each value of the parameter $\ell^{-1}$ chosen by the user.
\item Compute the multipoint Morisita index $I_{m=2,\ell^{-1}}$ for each value of the parameter $\ell^{-1}$ using Equation \ref{Eq_mindex}. Notice that the values of $I_{m=2,\ell^{-1}}$ are equal to those of $I_{m=2,\delta}$, since $\delta$ and $\ell^{-1}$ are two different ways of characterizing the size of the same cells. 
\item \label{item_lm} Carry out the linear regression of $\log{(I_{m=2,\ell^{-1}})}$ on $\log{(\ell^{-1})}$. Then $S_{m=2}$ is simply the slope of the regression model.
\item Compute $M_{m=2}$ using Equation \ref{Eq_mindid}.
\end{enumerate}
The procedure is illustrated in Figure \ref{Fig_Mindid} for $E=2$. On the right, the two features $F_1$ and $F_2$ have been rescaled to the $[0,1]$ interval and a 2-dimensional grid is superimposed over the data points. The number of cells along each of the two axes of the grid is equal to $4$. This is the value of the parameter $\ell^{-1}$ which allows the user to control the grid resolution. The calculation of $I_{m=2,\ell^{-1}}$ was iterated four times ($R=4$) for $\ell^{-1} \in \lbrace 1,2,3,4\rbrace$, and the results were used to draw the log-log plot shown on the left of the figure. The dashed line represents the linear regression model of Step \ref{item_lm}. Its slope is the Morisita slope $S_2$. 

In the next sections, $M_2$ will be computed using the MINDID algorithm \cite{Go15} whose complexity is $\mathcal{O}(N*E*R)$. MINDID is part of the \textsf{R} package ``IDmining'' \cite{GoR16}. It is its high execution speed that allows the efficient implementation of the unsupervised feature selection technique introduced in this paper. For instance, MINDID is able to compute the ID of the butterfly data set (see Section \ref{sim_data}) in $0.07$ seconds (s), $0.17$ s and $0.94$ s for respectively $N=10^3,10^4,10^5$. The experiment was carried out in the \textsf{R} environment using an Intel Core i7-2600 CPU @ 3.40 GHz along with 16.0 GB of RAM under Windows 7.

\subsection{Discussion About the Parameter $\ell^{-1}$}\label{sub_par}
The values of the parameter $\ell^{-1}$ can be chosen by drawing the log-log plot relating $\log{(I_{m=2,\ell^{-1}})}$ to $\log{(\ell^{-1})}$ for a sufficiently large set of values of $\ell^{-1}$. Then the values to be retained for the ID estimation are those corresponding to the linear part of the plot. Here is  some guideline for drawing the plot: 
\begin{enumerate}[noitemsep]
\item The minimum value that $\ell^{-1}$ can take on is $1$.
\item The maximum value of $\ell^{-1}$ must ensure that one grid cell at least contains two points. If all the occupied cells contain no more than one point, the value of the multipoint Morisita index falls down to zero and does not reflect the scaling behaviour of the data anymore. 
\item Between the minimum and the maximum, $\ell^{-1}$ can take on any integer value. In general, it follows a geometric sequence of ratio $1$ or $2$.
\end{enumerate}
Once the plot has been drawn, only the values of $\ell^{-1}$ consistent with a linear regression model must be kept for the ID estimation. Although a linear dependency is possible only for data sets exhibiting self-similarity over several scales, it is hardly ever a limitation, since many real world data sets follow such a behaviour \cite{Sousa07}.

It was also shown that the range of selected scales does not need to be very large in order to achieve good results in terms of redundancy detection and quantification \cite{Go15Esann,Go15}. However, in some extreme cases where $N$ is very low with regard to $E$, it may happen that the maximum value of $\ell^{-1}$ cannot be greater than $1$, making it impossible to compute the data ID. But this limitation does not prevent the application of the Morisita estimator to challenging case studies. And it is also worth mentioning that $M_2$ is more robust to small sample sizes than R\'{e}nyi's dimension of order $2$, $D_2$, that is used in most ID-based methods of data mining \cite{Go15}. 

\begin{figure}
\centering
\includegraphics[width=\linewidth]{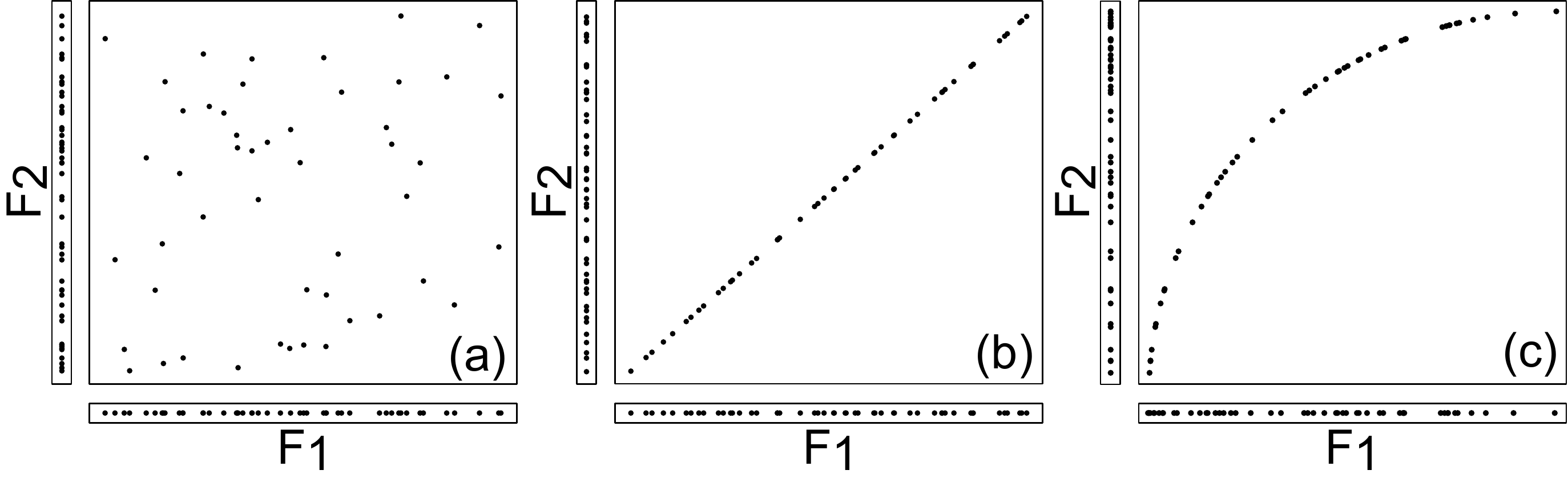}
\caption{Illustration of the three scenarios used to describe the possible redundancies between two variables (or features). Typically, two variables, $F_1$ and $F_2$, can be either (a) not redundant, (b) linearly redundant or (c) non-linearly redundant. In addition, the marginal point distributions are provided for the three scenarios.}\label{Fig_IDRedundancy}
\end{figure}

\section{Intrinsic Dimension and Redundancy}\label{ID_cor}
This section focuses on the use of the data ID to find redundancies among variables (or features). The case with only two variables is first presented and then extended to multivariate interactions.

Three scenarios are used to summarize the possible redundancies that can exist between two random variables. Each scenario considers 50 sampled data points and is illustrated in one of the panels of Figure \ref{Fig_IDRedundancy}. In the left-hand panel, two variables, called $F_1$ and $F_2$, are \textit{not redundant} and, with $ID(\cdot)$ denoting the ID of a data set, one can write that: 
\begin{equation}\label{equ_ID_1} 
ID(F_1,F_2)\approx ID(F_1)+ID(F_2)\approx 1+1=2.
\end{equation}This result follows from the fact that the data points cover both the 1-dimensional spaces constructed with the two variables taken separately and the 2-dimensional space of the two variables considered together. In contrast, if $F_1$ and $F_2$ are \textit{linearly} or \textit{non-linearly redundant}, as illustrated in the other two panels, their joint ID is approximately equal to $1$:
\begin{equation}\label{equ_ID_2}  
ID(F_1,F_2)\approx ID(F_1)\approx ID(F_2)\approx 1.
\end{equation}The resulting point patterns do not cover the 2-dimensional space. Instead, they cover a line in a similar way to what can be observed when $F_1$ and $F_2$ are taken separately. In the third scenario, the non-linear redundancy implies that the marginal distributions of the two features are not identical to the point distribution on the arc. However, it has only a small impact on the ID estimates as reported in \cite{Trai00}. This is the reason why ID-based estimators can effectively deal with non-linear redundancies. 

In other words, a feature (or variable) that is redundant does not (or hardly) contribute to increasing the ID of a data set. For instance, in the last two scenarios, one of the features is redundant with the other and can be disregarded without any loss of information. As a consequence, the data ID remains approximately unchanged after the addition of $F_2$ to the data set consisting solely of $F_1$ (and vice versa). In contrast, the first scenario requires that the two features be kept (i.e. none of them is redundant), and the data ID increases from $1$ to $2$ when $F_2$ is added to the data set consisting of $F_1$ (and vice versa).

The same reasoning applies to multivariate redundancy. Let $F_1$, $\ldots$, $F_k$ be $k$ random variables. Then, in the case where $F_k$ is not redundant with the $k-1$ other variables, Equation \ref{equ_ID_1} can be generalized as follows: 
\begin{equation}\label{equ_IDRed_1} 
ID(F_1,\ldots ,F_{k-1},F_k)\approx ID(F_1,\ldots ,F_{k-1})+ID(F_k)
\end{equation}and, in the case where $F_k$ is completely redundant, the multivariate version of Equation \ref{equ_ID_2} can be written as: 
\begin{equation}\label{equ_IDRed_2}  
ID(F_1,\ldots ,F_{k-1},F_k)\approx ID(F_1,\ldots ,F_{k-1}).
\end{equation}Finally, in the case where $F_k$ is only partially redundant, and according to the extension of ID to fractal dimensions (see e.g. Traina et al. \cite{Trai10} and De Sousa et al. \cite{Sousa07}), one can write that: 
\begin{equation}\label{equ_IDRed_3}  
ID(F_1,\ldots ,F_{k-1},F_k)-ID(F_1,\ldots ,F_{k-1})\approx W
\end{equation}where $W \in\left] 0,ID(F_k)\right[$ and where the less redundant $F_k$ is, the greater $W$. 

In the next subsection, the proposed algorithm executes a procedure of feature selection which follows directly from Equations \ref{equ_IDRed_1}, \ref{equ_IDRed_2} and \ref{equ_IDRed_3}. However, in real world applications, variable distributions can often greatly depart from the simple ones shown in Figure \ref{Fig_IDRedundancy}. In such situations, Equations \ref{equ_IDRed_1}, \ref{equ_IDRed_2} and \ref{equ_IDRed_3} remain unchanged, but
it must be noticed that:
\begin{equation}
ID(F_i)\leqslant 1 \quad \forall i\in\lbrace 1,2,\ldots,k\rbrace
\end{equation}where the equality holds only if the variables are uniformly distributed. Thus, for Equations \ref{equ_IDRed_1}, \ref{equ_IDRed_2} and \ref{equ_IDRed_3}, one can specify that:
\begin{equation}\label{equ_non_unif} 
ID(F_1,\ldots ,F_{k-1},F_k)\leq k.
\end{equation}

\begin{algorithm}[t]
\caption{MBRM}\label{mbrm_algo}
\textbf{INPUT:} 

A dataset $A$ with $E$ features $F_{1,\ldots, E}$.
 
A vector $L$ of values $\ell^{-1}$.
 
An integer $C$ ($\leq E$) indicating the number of steps of the SFS procedure to be performed (by default $C=E$).
 
Two empty vectors of length $C$: $SelF$ and $IDF$ for storing, respectively, the names of the selected features and the data ID estimates.
 
An empty matrix $Z$ for storing the selected features.

Optional: the ID estimate $IDA$ of the full data set $A$.

\textbf{OUTPUT:} $SelF$ and $IDF$. 
\begin{algorithmic}[1]
\STATE Rescale each feature to $[0,1]$.
\STATE Unless $IDA$ is given: $IDA=M_2(A)$ (MINDID used with $L$)
\FOR{$i = 1 \ \TO \ C$}
\FOR{$j = 1 \ \TO \ (E+1-i)$}
\STATE $Diff(Z,F_j)=\left| IDA-M_2(Z,F_j)\right|$ (MINDID used with $L$)
\ENDFOR
\STATE Store in $SelF[i]$ the name of the $F_j$ yielding the lowest value of $Diff$. 
\STATE Store the corresponding value of $M_2(Z,F_j)$ in $IDF[i]$. 
\STATE Remove the corresponding $F_j$ from $A$ and add it into $Z$.
\ENDFOR\\
\end{algorithmic}
\end{algorithm}

\section{The Morisita-Based Filter for Redundancy Minimization}\label{ID_RM}
The Morisita-Based filter for Redundancy Minimization (MBRM) aims to select the smallest subset of features necessary to carry all the information content of a data set. MBRM follows a SFS search strategy and, in each step of the procedure, it searches for the feature which yields the greatest value of $W$ (i.e. the feature which carries the largest amount of new information according to Equation \ref{equ_IDRed_3}). To achieve this goal, a fast solution, robust to sample size and noise, is to reduce the following difference to zero (see Algorithm \ref{mbrm_algo}):
\begin{equation}\label{equ_MBRM} 
Diff(F):=\left| M_2(A)-M_2(F)\right|
\end{equation}where $A$ is a data set consisting of $E$ features $F_{1,\ldots, E}$, $F$ is a set of $i$ features of $A$ with $i\leqslant E$ and $M_2(\cdot)$ denotes the estimation of the data ID using the Morisita estimator $M_2$. More precisely, MBRM starts with $i=1$ and searches for the individual feature which contributes the most to reducing $Diff$. Once identified, this feature is retained. Then the algorithm carries on in searching for the second feature ($i=2$) which leads to the largest possible decrease in $Diff$ when combined with the previously retained feature. The operation is iterated until $i=C$ with $C\leqslant E$. Typically, $C$ can be set to a value lower than $E$ (by default $C=E$) if it is assumed that $A$ is highly affected by redundancy. Finally, the set of features contributing to decreasing the value of $Diff$ to approximately $0$ (i.e. $Diff\approx 0$) is the smallest set of features carrying all the information content of $A$.

\begin{figure}[t]
\centering
\includegraphics[width=\linewidth]{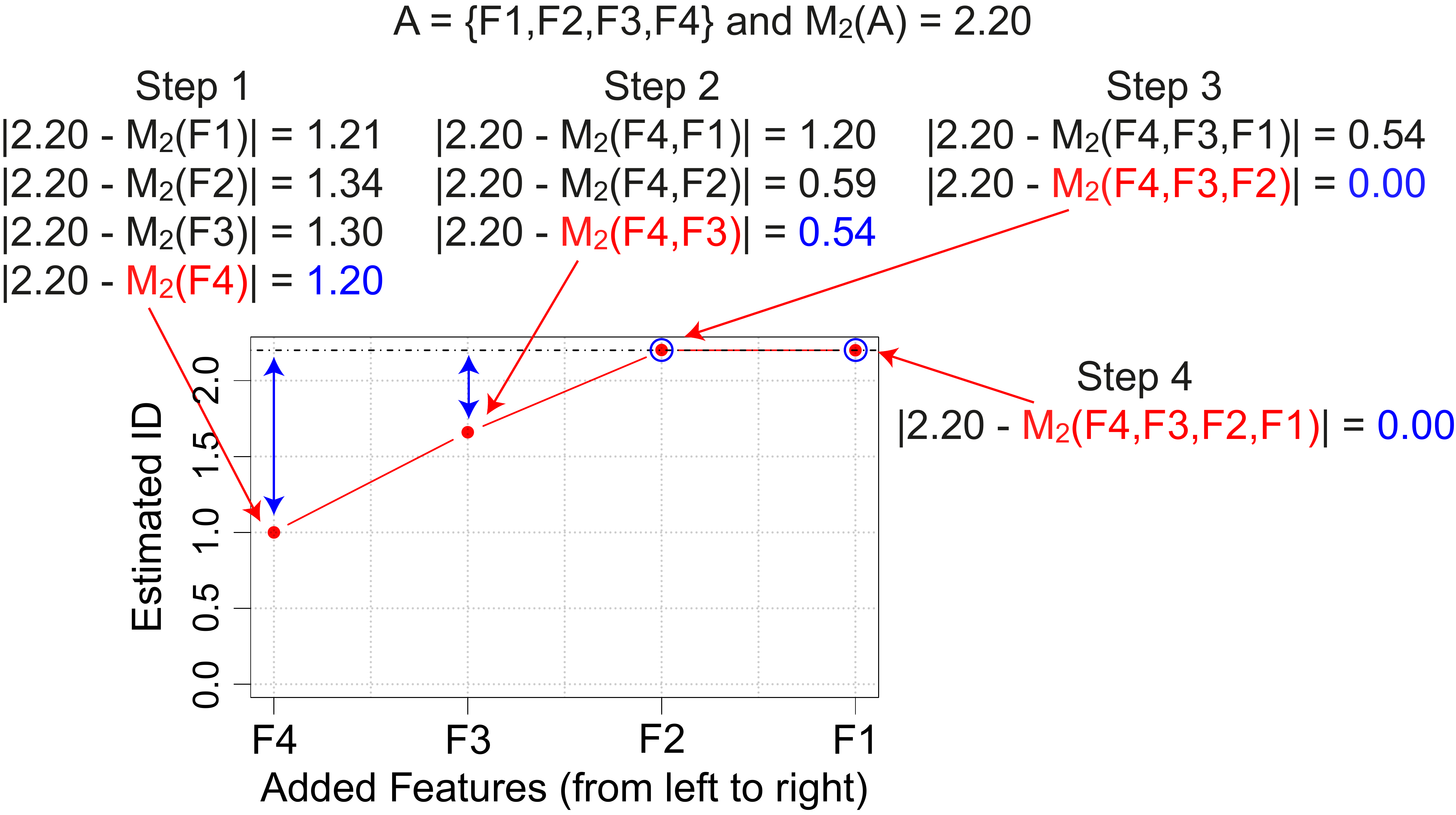}
\caption{Illustration of how MBRM works. A data set $A$ is used. It consists of four features $F_1$, $F_2$, $F_3$ and $F_4$. The result shows that $F_1$ is redundant and the three other features are enough to carry all the information content of $A$.}\label{Fig_MBRM}
\end{figure}

Figure \ref{Fig_MBRM} illustrates how MBRM works. In this example, the data set $A$ consists of four features, namely $F_1$, $F_2$, $F_3$ and $F_4$, and the full data ID is equal to $2.20$. In Step 1 ($i=1$), the feature that contributes the most to reducing $Diff$ is $F_4$, and the value of $M_2(F_4)$ is reported in the plot. Then MBRM moves on to Step 2 ($i=2$) where $F_3$ is identified as the feature leading to the largest decrease in the value of $Diff$ when combined with $F_1$. Therefore, the value of $M_2(F_4,F_3)$ is added to the plot in second position. It appears before the values of $M_2(F_4,F_3,F_2)$ and $M_2(F_4,F_3,F_2,F_1)$ which result from Steps 3 ($i=3$) and 4 ($i=4$). Eventually, the plot highlights that only $F_4$, $F_3$ and $F_2$ contribute to increasing the ID estimates of the data. Hence, it can be concluded that the removal of $F_1$ will have no impact on the information content of $A$. In this simple example, if $C$ had been set to $3$, MBRM would have returned only $F_4$, $F_3$ and $F_2$. No information would have been lost as indicated by the equality between $M_2(F_4,F_3,F_2)$ and $M_2(A)$.

MBRM relies on the recent MINDID algorithm \cite{Go15} for the computation of $M_2$ (see Section \ref{Mindex}). Consequently, MBRM takes as input a vector $L$ consisting of $R$ values of $\ell^{-1}$. It will be shown in the next section that these values can be chosen with regard to the full data set $A$ and remain unchanged throughout the steps of the sequential forward search. In the same way as MINDID, MBRM is linear on both the number $N$ of points and the number $R$ of scales. But the complexity of the sequential search is quadratic on the number $E$ of features. In spite of this limitation, the execution time remains competitive thanks to the fast computation speed of MINDID. It can also be significantly reduced by setting $C$ to a low value. For instance, when applied to the butterfly data set (see Section \ref{sim_data}), MBRM runs in $2.14$ seconds (s), $3.06$ s and $12.45$ s for respectively $N=10^3,10^4,10^5$ and $C=E$. And, with $C=3$ (i.e. the number of features necessary to carry all the information content of the data), the execution time of MBRM is reduced to $0.68$ s, $1.04$ s and $4.07$ s for the same numbers of points. The computations were performed in the \textsf{R} environment using an Intel Core i7-2600 CPU @ 3.40 GHz along with 16.0 GB of RAM under Windows 7.

Finally, a possible extension of MBRM would be to allow the user to provide the value of the full data ID in cases where it is known a priori. As a result, MBRM would never need to be applied to spaces with more than $C$ dimensions. However, in this paper, the full data ID will always be estimated using $M_2(\cdot)$ within MBRM, since it is hardly ever available in advance.

\section{Experimental Study}\label{exp_res}
In this section, numerical experiments designed to test the MBRM algorithm are presented. A simulated data set (see Subsection \ref{sim_data}) was used to examine the impact of sample size and noise as well as the ability of the algorithm to capture non-linear relationships between features. Monte-Carlo simulations were also performed in order to quantify the variability of the results. In addition, MBRM was applied to real world case studies (see Subsection \ref{real_data}). In this context, the preserved amount of relevant information was assessed by comparisons with benchmark feature selection techniques through the use of Random Forests \cite{Bre01}. Notice that all the experiments were carried out in the \textsf{R} environment \cite{Rcran,GoR16}.

\begin{figure}[t]
\centering
\includegraphics[width=\linewidth]{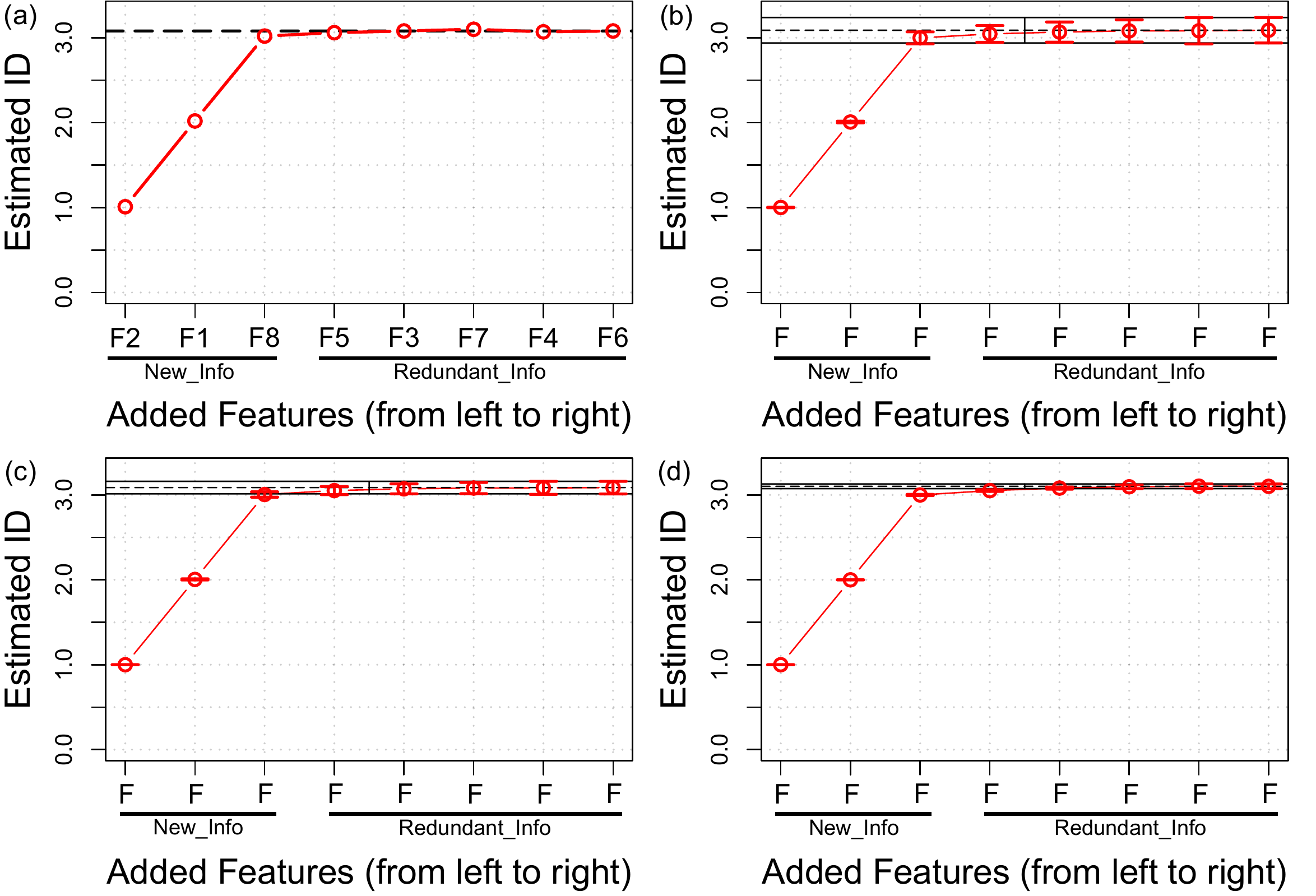}
\caption{MBRM applied to the input variables of the butterfly data set: (a) $1$ simulation with $N=1000$, (b) $100$ simulations with $N=1000$, (c) $100$ simulations with $N=2000$, (d) $100$ simulations with $N=10000$. In each panel, the dashed line indicates the ID of the full data set and, when $100$ simulations are used, the mean ID estimates are given along with error bars indicating $\pm$ the standard deviations. Notice that the names of the features were shortened to $F$ because the way they are ordered by the SFS search strategy can change between the simulations.}\label{Fig_sampsize}
\end{figure}

\subsection{Simulated Data Set}\label{sim_data}
The MBRM algorithm was applied to the input space of the butterfly data set \cite{Go15Esann}. It consists of eight variables (or features): two uniformly distributed variables $F_1,F_2 \in \left] -5,5\right[$, $F_3=log_{10}{(F_1+5)}$, $F_4 = F_1^2-F_2^2$, $F_5 = F_1^4-F_2^4$, a third uniformly distributed variable $F_6\in \left]-5,5\right[$, $F_7=log_{10}{(F_6+5)}$ and $F_8=F_6+F_7$. The butterfly data set is then generated by random sampling of $F_1$, $F_2$ and $F_6$. In this paper, the following sample sizes were considered: $N=1000,2000,10000$, and for each of them, $100$ versions of the data set were produced.

The MBRM algorithm was applied to the butterfly data set for each of the above-mentioned sample size. The results are displayed in Figure \ref{Fig_sampsize}. The first panel shows the plot provided by MBRM when applied to one simulation of $1000$ data points. After the addition of $F_8$ to $F_2$ and $F_1$, the ID estimate is equal to $3.02$ (i.e. $M_2(F_2,F_1,F_8)=3.02$) and a clear cut-off point can be observed, since the remaining features hardly contribute to increasing the value of the full data ID (i.e. hardly contribute to reducing the value of $Diff$ defined in Equation \ref{equ_MBRM}). Consequently, the MBRM algorithm has detected that only three features, namely $F_1$, $F_2$ and $F_8$, are enough to carry all the information content of the butterfly data set. This is correct by construction, since the full data set was generated from $F_1$, $F_2$ and $F_6$ and since $F_8$ is redundant with $F_6$. MBRM has thus successfully fulfilled its goal.

The remaining panels focus on the variability of the ID estimates with regard to the number of data points. The plots depict the evolution of the mean ID estimates during the SFS search procedure, and the errors bars represent $\pm$ the standard deviations of the estimates. The mean value and the standard deviation of the full data ID estimates are also given. They are denoted respectively by the dashed and solid black lines. As expected, the variability tends to be higher for lower sample sizes. This is clearly visible for the full data ID estimates, but it never prevents MBRM from detecting that only three features can convey all the information content of the butterfly data set. Notice, however, that the ordering of the features resulting from the SFS search procedure can change from one simulation to the next. Consequently, the names of the features were shortened to $F$ in each panel.

\begin{table}
\centering 
\footnotesize
\begin{tabular}{cc}
\hline 
$N$    &  First Three Features (Occurrences)\\ 
\hline 
\hline
1000   & $F_1$,$F_2$,$F_6$ ($72$); $F_1$,$F_2$,$F_8$ ($19$); $F_1$,$F_2$,$F_7$ ($7$); $F_1$,$F_3$,$F_6$ ($1$); $F_1$,$F_4$,$F_6$ ($1$)\\   
2000   & $F_1$,$F_2$,$F_6$ ($82$); $F_1$,$F_2$,$F_8$ ($18$)\\ 
10000  & $F_1$,$F_2$,$F_6$ ($100$)                         \\ 
\hline 
\end{tabular}
\caption{The first three features selected by MBRM (not necessarily in that order) when applied successively to $100$ simulations of the butterfly dataset for $3$ different sample sizes. The number of occurrences of each triplet is indicated in brackets.}
\label{Tab_Feat_Order}
\end{table} 

Another interesting point is the composition of the set of features picked first by MBRM. By construction, three features should be identified by the algorithm as being necessary to preserve the information content of the butterfly data set. Several triplets of features could work, but MBRM favours $F_1$, $F_2$ and $F_6$ as indicated in Table \ref{Tab_Feat_Order}. This is mainly due to the non-linear construction of $F_3$, $F_4$, $F_5$ and $F_7$ (because non-linearity influences the way the data points are distributed over the data manifold as shown in the right-hand panel of Figure \ref{Fig_IDRedundancy}, which can in turn have a slight effect on the ID estimates) and to the fact that $F_4$ and $F_5$ are fully redundant with $F_1$ and $F_2$ considered jointly rather than independently. For these reasons, when the sample size is large enough, the features constructed to be redundant do not maximize the ID estimates as much as $F_1$, $F_2$ or $F_6$ and, consequently, they are not picked first by MBRM.

\begin{figure}[t]
\centering
\includegraphics[width=\linewidth]{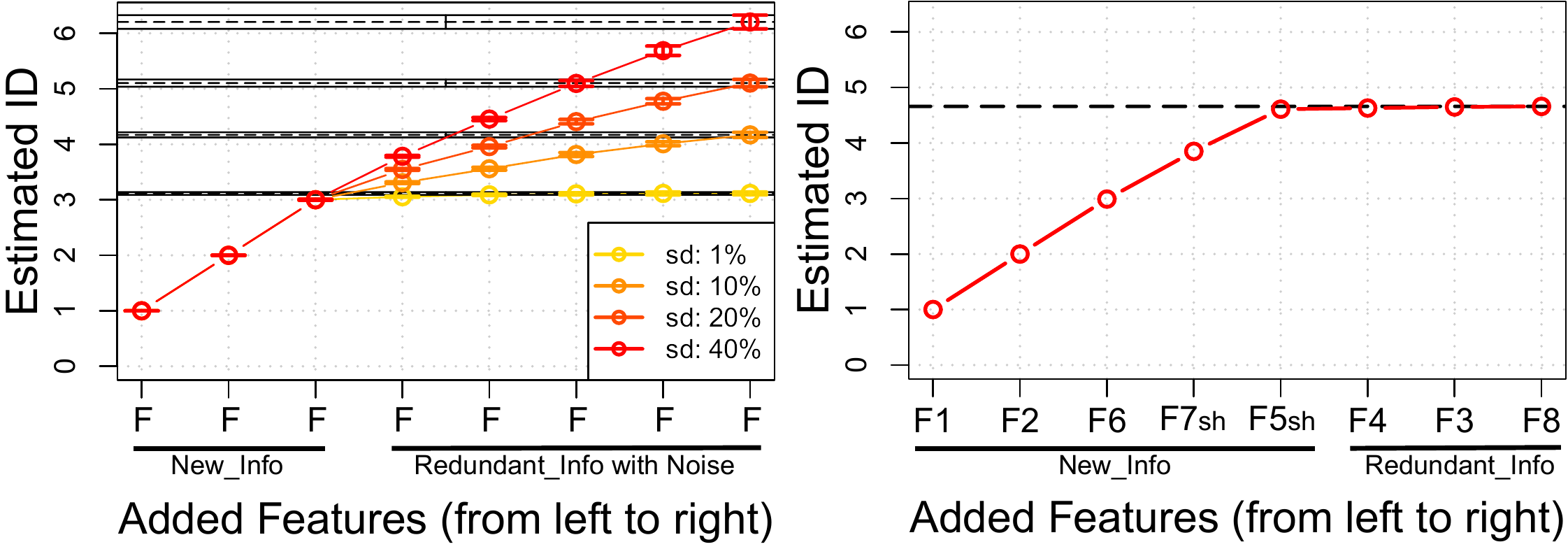}
\caption{MBRM applied to the input variables of the butterfly data set with $N=10000$: (left) $100$ simulations with different levels of Gaussian noise, (right) $1$ simulation after the shuffling of $F_5$ and $F_7$. The dashed lines indicate the ID of the full data sets, and in the left-hand panel, the mean ID estimates are given along with error bars indicating $\pm$ the standard deviations.}\label{Fig_noise_shuf}
\end{figure}

A good algorithm should also be robust to noise. Consequently, another experiment was carried out for which $F_3$, $F_4$, $F_5$, $F_7$ and $F_8$ were corrupted with a Gaussian noise. For each of these features, the mean of the noise was fixed at $0$, and the standard deviation (sd) was successively set to $1\%$, $10\%$, $20\%$ and $40\%$ of the standard deviation of the original feature. The left-hand panel of Figure \ref{Fig_noise_shuf} shows the results. At $1\%$, the noise hardly affects MBRM, and the cut-off point is visible up to $20\%$. Beyond this threshold, the algorithm has difficulty fulfilling its task, but it is quite normal, since it could be argued that a data set affected by more than $40\%$ of noise should be used with caution. Moreover, it is also clear that the full data ID and the variability of the estimates increase with the level of noise. This result is simply a consequence of the fact that the noise enlarges the portion of space occupied by the data points. 

Finally, the right-hand panel of Figure \ref{Fig_noise_shuf} displays the results provided by MBRM after the shuffling of $F_5$ and $F_7$. The shuffling makes the two features completely independent of, respectively, $F_1$ and $F_6$ (i.e non-redundant with $F_1$ and $F_6$) and, consequently, MBRM detects that they carry new information, which is correct. Notice also that the full data ID is lower than $5$, as expected from Equation \ref{equ_non_unif}: since $F_5$ and $F_7$ are not uniformly distributed, the data ID is lower than $k=5$. This means that, in general, the value of the full data ID cannot be used as the number of features to be retained. It must only be considered a lower bound. And, of course, selecting less features than the value of the ID will always lead to a loss of information. For instance, $M_2(F_1,F_2,F_4)\approx 2$, which implies that two features, say $F_1$ and $F_2$, are necessary to correctly synthesize the data, but $F_4$ alone cannot account for all the available information. Differently put, it is possible to reconstruct $F_4$ from $F_1$ and $F_2$, but neither $F_1$ nor $F_2$ can be perfectly retrieved from $F_4$.

\subsection{Real World Case Studies}\label{real_data}
\begin{figure}
\centering
\includegraphics[width=\linewidth]{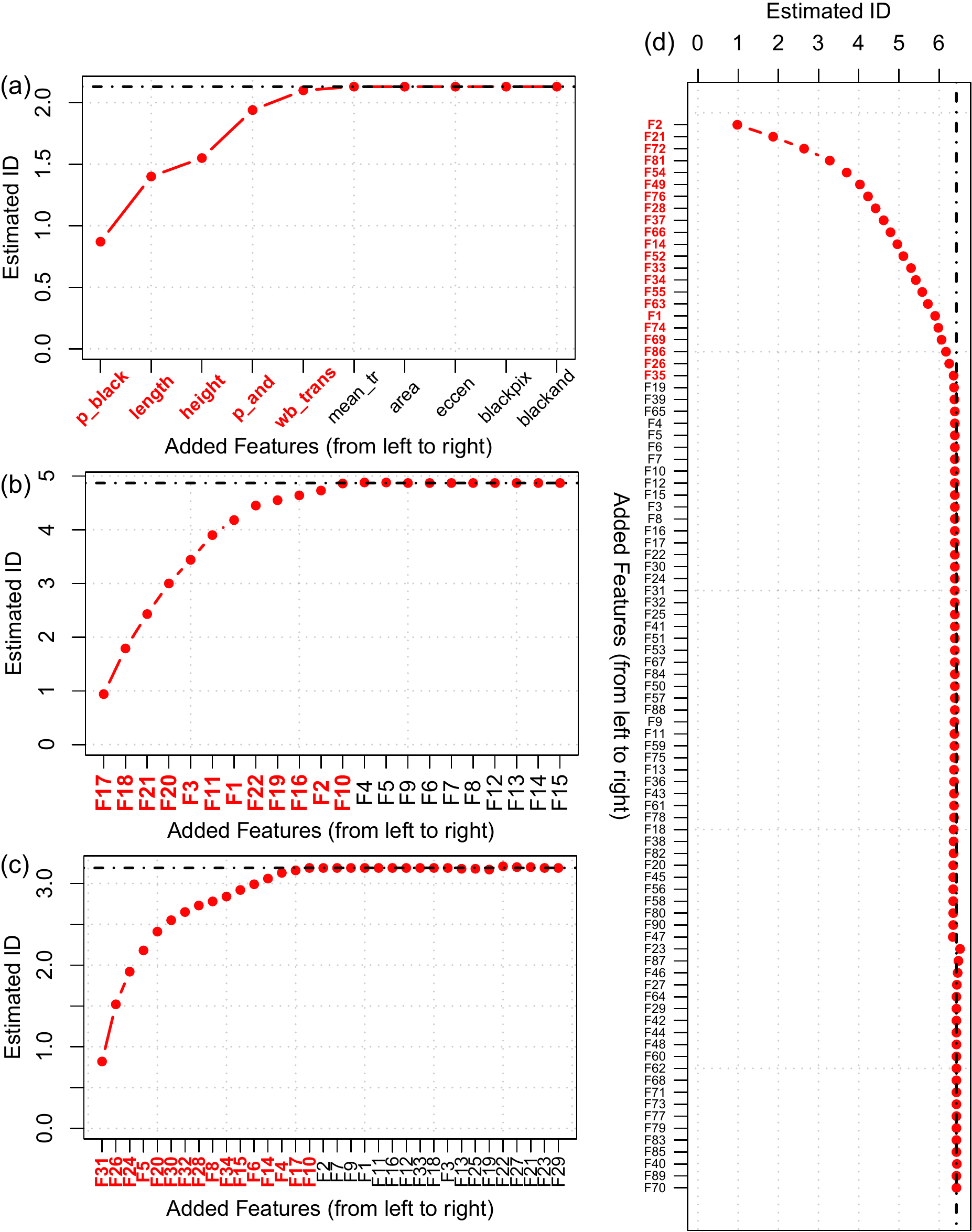}
\caption{Results of MBRM: (a) PageBlocks, (b) Parkinson, (c) Ionoshpere, (d) LIBRAS Movements. The features to be selected are indicated in bold red, and the dashed lines correspond to the ID of the full data sets.}\label{Fig_realdata}
\end{figure}

Four real world data sets from the UCI machine learning repository \cite{Lich} were used: PageBlocks, Parkinson, Ionosphere and LIBRAS movements. Duplicate data points were removed, and the main characteristics of the resulting data are briefly summarized in Table \ref{Tab_data}. In addition, the last column of the table provides the values of the parameter $\ell^{-1}$ used to apply the MBRM algorithm to the input space of the data sets. The complete feature selection procedure was performed as follows (for each data set):
\begin{enumerate}[noitemsep]
\item The plot relating $\log{(I_{m=2,\ell^{-1}})}$ and $\log{(\ell^{-1})}$ was drawn for the $E$ features taken together and for values of $\ell^{-1}$ ranging from $1$ to the highest possible value (i.e. the value ensuring the presence of two points in at least one of the $Q$ cells).
\item The range of values of $\ell^{-1}$ corresponding to the linear part of the log-log plot was retained. For each of the four data sets, this final range was that of step $1$.
\item If the upper bound of the range was lower than $30$, each integer value of the range was used in the final set given in Table \ref{Tab_data}. If not, for efficiency purposes, only the values following a geometric sequence of ratio $2$ were retained. 
\item MBRM was applied to each data set with the parameter values of Step $3$, and the features necessary to approximately reach the full data ID were selected.
\end{enumerate}
The results are given in Figure \ref{Fig_realdata} and summarized in Table \ref{Tab_res_MBRM}. For each data set, MBRM offers a clear cut-off point allowing an easy distinction between the features to be selected and those considered redundant. Moreover, it leads to a significant reduction in the data dimensionality. In the case of LIBRAS, only $24\%$ of the original features were selected, and the other data sets were reduced by half.

\begin{table}
\centering 
\footnotesize
\begin{tabular}{ccccc}
\hline 
Data Sets   & $N$    &  $E$  &  \# Classes & Parameter $\ell^{-1}\in$\\ 
\hline 
\hline
PageBlocks  & $5393$ & $10$  &  $5$        & $\lbrace 2^n\mid n=0,1,2,\ldots ,11\rbrace$\\   
Parkinson   & $195$  & $22$  &  $2$        & $\lbrace 1,2,3,\ldots ,8\rbrace$           \\
Ionosphere  & $350$  & $34$  &  $2$        & $\lbrace 1,2,3,\ldots ,13\rbrace$          \\  
LIBRAS      & $330$  & $90$  &  $15$       & $\lbrace 1,2,3,4,5\rbrace$                 \\
\hline 
\end{tabular}
\caption{Data set characteristics and values of the parameter $\ell^{-1}$.}
\label{Tab_data}
\end{table} 

\begin{table}
\centering 
\footnotesize
\begin{tabular}{cccc}
\hline 
Data Sets   & \# Sel. Feat.& $M_2(All \ Feat.)$ & Execution Time (s)\\ 
\hline 
\hline
PageBlocks  & $5 \ (50\%)$   &       $2.13$     &     $3.45$        \\
Parkinson   & $12 \ (52\%)$  &       $4.87$     &     $5.32$        \\    
Ionosphere  & $16 \ (47\%)$  &       $3.19$     &     $25.68$       \\ 
LIBRAS      & $22 \ (24\%)$  &       $6.43$     &     $130.81$      \\
\hline 
\end{tabular}
\caption{Summary of the results provided by MBRM and execution time (in seconds). In the column ``\# Sel. Feat.'', the values in brackets are the percentages of selected features with regard to the total numbers of features.}
\label{Tab_res_MBRM}
\end{table} 

In terms of efficiency, MBRM is linear on $N$, but its bottleneck is the number of features. Therefore, it takes $3.45$ seconds (s) to run the algorithm on PageBlocks (using \textsf{R} and an Intel Core i7-2600 CPU @ 3.40 GHz along with 16.0 GB of RAM under Windows 7), while $130.81$ s are necessary for LIBRAS. Nevertheless, if it can be assumed that a data set contains many redundant features, the parameter $C$ can be set to a value lower than $E$ to shorten the execution time. For instance, for the LIBRAS data set, if $C$ is set to the number of selected features, the execution time of MBRM is cut down to $17.69$ s.

The effectiveness of MBRM in fulfilling redundancy minimization must also be assessed. In the case of the PageBlocks data set, it is straightforward, since the five features considered redundant by the algorithm are truly redundant. In other words, they can be computed from the selected features as explained on the UCI machine learning repository \cite{Lich}. Regarding the remaining data sets, no a priori knowledge about the redundancy between the features is available, and a more complex procedure is required to evaluate the amount of information kept by the selected features. A possible solution is to use an algorithm, such as Random Forest (RF), to classify the data points according to the output variables. In this paper, the choice of RF was motivated by its recently recognized benchmark performance and its reputation of working well in high-dimensional spaces \cite{Bre01,Has09,Zh16}. Then the fundamental idea consists in performing the classification task twice, once with all the input features and once with only the selected features. If the reduced subset does not lead to a higher error rate than that obtained with the full data set, the relevant information has been preserved, and the feature selection can be considered successful.

The same idea can also be extended to compare different techniques of feature selection. A technique that leads to a significant increase in the classification error does not perform as well as a technique that maintains the accuracy obtained with the full data set. Here, four benchmark algorithms were used: Fractal Dimension Reduction (FDR) \cite{Trai00}, Feature Similarity Feature Selection (FSFS) based on the maximum information compression index \cite{Mit02}, Laplacian Score (LScore) \cite{He06} and Multi-Cluster Feature Selection (MCFS) \cite{Cai10}. For FSFS, LScore and MCFS, the number of selected features was set to the number of features selected by MBRM, and the number of nearest neighbours, $k$, of LScore and MCFS was fixed at $5$. Finally, regarding MCFS, the number of used eigenvectors was set to the number of classes.

The exact evaluation procedure implemented in this research was repeated six times for each data set: once with all the features, once with the features selected by MBRM and four more times with the features selected by FDR, FSFS, LScore and MCFS. This procedure follows four steps and is inspired by similar procedures suggested in \cite{Ja97,Reu03}:\begin{enumerate}[noitemsep]
\item $20\%$ of the data points were randomly assigned to a test set and the remaining ones were passed on to Step 2.
\item 10-fold cross-validation was performed to tune the two parameters of RF (these two parameters are the number of trees and the number of predictors processed at each split of the trees). The Overall Accuracy (OA) was used as evaluation metric.
\item \label{item_tst}A RF model was trained with the values of the two parameters of Step 2 and was then used to classify the data points of the test set. Finally, the test OA, $OA_{t}$, was computed using the following formula:\begin{equation}\label{Eq_OA} 
OA_{t}=\frac{1}{N_t} \sum_{i=1}^{N_t}I(y_i = \hat{y_i})
\end{equation}where $I(y_i = \hat{y_i})$ is an indicator variable, $N_t$ is the number of data points in the test set, $y_i$ refers to the true class label of the $i^{th}$ observation and $\hat{y_i}$ is the predicted class label.
\item Steps 1 to 3 were repeated $20$ times, and the mean test OA (in percent) along with the corresponding standard deviation were computed. These two values are given in Table \ref{Tab_res_RF_OA} for each data set.
\end{enumerate}

Table \ref{Tab_res_RF_OA} shows that the mean overall accuracies obtained using the features selected by MBRM are better than or close to those computed from the full data sets. By ``close'', it is meant that they are within half a standard deviation of one another. This suggests that the selected features carry all the information content relevant to the classification tasks and that the MBRM algorithm has fulfilled its purpose. Moreover, compared to the four benchmark techniques used in the experiments, MBRM always provided equal or better results. For instance, in the case of the LIBRAS data set, LScore, FSFS and MCFS led to a significant increase in the mean overall accuracy, while MBRM maintained the value obtained using all the features. Regarding FDR, the algorithm could not be run because the log-log plot necessary to estimate the ID did not exhibit any linear behaviour. The same remark holds for the Ionosphere data. 

\begin{landscape}
\begin{table}
\centering 
\footnotesize
\begin{tabular}{cccccc|c}
\hline 
Data Sets   &      MBRM      &       FDR       &      FSFS       &     LScore       & MCFS &    All Feat.  \\ 
\hline 
\hline
PageBlocks & {\boldmath$97.37 \ (0.51)$} & $97.14 \ (0.52)$ & $97.09 \ (0.63)$ & $96.79 \ (0.53)$  & \underline{$97.36 \ (0.58)$} & $97.54\ (0.56)$ \\
Parkinson  & {\boldmath$92.05 \ (5.32)$} & $91.54 \ (5.46)$ & \underline{$91.92 \ (5.41)$} & $89.49 \ (4.77)$  & $90.13\ (3.93)$  & $90.38\ (5.93)$ \\    
Ionosphere & {\boldmath$91.93 \ (3.29)$} &        -         & $90.43 \ (2.94)$ & \underline{$91.29 \ (3.31)$}  & $91.21\ (3.08)$  & $92.57\ (2.92)$ \\ 
LIBRAS     & {\boldmath$76.29\ (3.82)$}  &        -         & $66.74 \ (5.91)$ & $57.80 \ (5.76)$  & \underline{$69.62 \ (5.81)$} & $75.68\ (4.68)$\\
\hline 
\end{tabular}
\caption{Mean overall accuracies (over $20$ random splits and in percent) yielded by Random Forests when applied to the features selected by MBRM, FDR, FSFS, LScore, MCFS and to the full data sets. The standard deviations are indicated in brackets. Besides, the best results provided by the feature selection techniques are shown in bold, and the second best results are underlined. The goal is to be higher than or (if smaller) as close as possible to the reference values given in the column ``All Feat.''.}
\label{Tab_res_RF_OA}
\end{table}

\begin{table}
\centering 
\footnotesize
\begin{tabular}{cccccc|c}
\hline 
Data Sets   &      MBRM      &       FDR       &      FSFS       &     LScore       & MCFS &    All Feat.  \\ 
\hline 
\hline
PageBlocks &{\boldmath$84.53 \ (2.44)$} & $83.04 \ (2.53)$  & $82.78 \ (3.18)$  & $80.73 \ (2.38)$  & \underline{$84.41 \ (3.01)$}  & $85.56\ (2.95)$ \\
Parkinson  &{\boldmath$77.06 \ (15.15)$}& $75.40 \ (15.84)$ & \underline{$76.54 \ (15.70)$} & $71.13 \ (13.27)$ & $72.41 \ (10.20)$ & $71.94\ (18.55)$ \\    
Ionosphere &{\boldmath$82.36 \ (7.10)$} &        -          & $78.91 \ (6.64)$  & \underline{$81.02 \ (6.79)$}  & $80.68 \ (6.87)$  & $83.84\ (6.16)$ \\ 
LIBRAS     &{\boldmath$74.40 \ (4.11)$} &        -          & $64.12 \ (6.33)$  & $54.54 \ (6.29)$  & \underline{$67.24 \ (6.16)$}  & $73.73\ (5.04)$\\
\hline 
\end{tabular}
\caption{Mean Kappa coefficients (over $20$ random splits and multiplied by $100$) yielded by Random Forests when applied to the features selected by MBRM, FDR, FSFS, LScore, MCFS and to the full data sets. The standard deviations are indicated in brackets. Besides, the best results provided by the feature selection techniques are shown in bold, and the second best results are underlined. The goal is to be higher than or (if smaller) as close as possible to the reference values given in the column ``All Feat.''.}
\label{Tab_res_RF_K}
\end{table}
\end{landscape}

In addition to the overall accuracy, Cohen's Kappa coefficient of agreement $\kappa$ \cite{Coh60} was also computed in Step \ref{item_tst} of the evaluation procedure (see Table \ref{Tab_res_RF_K}). Cohen's kappa is an evaluation metric 
that takes into account the number of correctly classified data points that may occur by chance. This way, the possible bias of the OA towards large classes is reduced. Cohen's Kappa is commonly used in data mining (see e.g. \cite{Ste16,Vo14}), and it is given by:\begin{equation}\label{Eq_K} 
\kappa_{t}=\frac{N_t\sum_{c=1}^{B}T_c-\sum_{c=1}^{B}G_cP_c}{N_t^2-\sum_{c=1}^{B}G_cP_c}
\end{equation}where the subscript $t$ indicates that the coefficient is computed on a test set, $B$ is the number of classes, $T_c$ indicates the number of correctly classified samples for class $c$ and $N_t$ is the number of data points in the test set. Finally, $G_c$ and $P_c$ are the actual number of samples belonging to class $c$ and the number of samples classified in this class. Cohen's Kappa ranges in $[-1,1]$, but negative values are hardly ever met. It is equal to $1$ in case of complete agreement and to zero (or below) if a classifier does not perform better than what would be expected from pure randomness. The results are provided in Table \ref{Tab_res_RF_K} and confirm those of Table \ref{Tab_res_RF_OA}. Notice, however, that they are lower and more prone to variability than the results yielded by the overall accuracy. But this is consistent with the way Cohen's Kappa accounts for the agreement due to chance.

Based on the numerical experiments presented in this section, it can be concluded that MBRM performed well and that it is a new promising tool for redundancy minimization.

\section{Conclusion}
This paper introduces a new algorithm for unsupervised feature selection called Morisita-Based filter for Redundancy Minimization (MBRM). MBRM relies on the Morisita estimator of Intrinsic Dimension (ID) and aims to identify the smallest subset of features containing all the information content of a data set. It was successfully tested on simulated data with different levels of sample size and noise. In addition, real world case studies from the UCI machine learning repository were used. MBRM turned out to be effective in a wide range of situations characterized by different numbers of data points, features and classes. When no information about the redundancy between the features was available in advance, a comprehensive procedure based on random forests was implemented to assess the performance of the feature selection. The classification results demonstrated that MBRM did not lead to any loss of relevant information, while it cuts down the size of every data set by half or more. Comparisons with benchmarks techniques confirmed the promising performance of the proposed algorithm. MBRM has also a couple of practical advantages over more traditional techniques of unsupervised feature selection. First, it is able to determine how many features should be kept. Second, the values of its parameter $\ell^{-1}$ can be set without resorting to any learning machine or a priori knowledge. Finally, it was shown that MBRM was able to outperform the FDR algorithm due to the Morisita estimator of ID. 

From a broader perspective, this research contributes to highlighting that the concept of ID can help mitigate issues raised by large data sets. Future research will be devoted to challenging applications in hyperspectral remote sensing. In this context, comparisons between ID-based and traditional methods will be carried out, and the use of ID to perform advanced data mining tasks other than feature selection will be thoroughly explored. In addition, the automatic parallelization of the inner loop of the MBRM algorithm will also be examined.

\section{Acknowledgements}
The authors are grateful to the anonymous reviewers for their helpful and constructive comments that contributed to improving the paper. They also would like to thank Michael Leuenberger and Mohamed Laib for many fruitful discussions about machine learning and statistics.

\bibliographystyle{elsarticle-num}
\bibliography{References}
 
\end{document}